\documentclass[10pt,twocolumn,letterpaper]{article}

\usepackage{iccv}
\usepackage{times}
\usepackage{epsfig}
\usepackage{graphicx}
\usepackage{amsmath}
\usepackage{amssymb}

\usepackage{threeparttable}
\usepackage{cite}
\usepackage{comment}
\usepackage{adjustbox}
\usepackage{textcomp}
\usepackage{array,multirow,graphicx}
\usepackage[table]{xcolor}
\usepackage{colortbl}
\usepackage{bbm}
\usepackage{makecell}
\usepackage[percent]{overpic}

\usepackage{booktabs}
\usepackage{arydshln}

\usepackage{pifont}
\newcommand{\cmark}{\ding{51}}
\newcommand{\xmark}{\ding{55}}

\usepackage[pagebackref=true,breaklinks=true,letterpaper=true,colorlinks,bookmarks=false]{hyperref}

\iccvfinalcopy %

\def\iccvPaperID{****} %

\def\eg{\emph{e.g}\onedot} 
\def\ie{\emph{i.e}\onedot} 
 
 \def\vs{\emph{vs}\onedot}
\def\wrt{w.r.t\onedot} 
\def\etal{\emph{et al}\onedot}

\begin{document}

\title{SPGNet: Semantic Prediction Guidance for Scene Parsing}

\author{Bowen Cheng$^{1}$, Liang-Chieh Chen, Yunchao Wei$^{1,3}$, Yukun Zhu, Zilong Huang$^1$,\\Jinjun Xiong$^2$, Thomas S. Huang$^{1}$, Wen-Mei Hwu$^1$, Honghui Shi$^{2,1,4}$\\
\\
{$^1$UIUC, $^2$IBM Research, $^3$ReLER, UTS, $^4$University of Oregon}}

\maketitle

\begin{abstract}
Multi-scale context module and single-stage encoder-decoder structure are commonly employed for semantic segmentation. Multi-scale context module aggregates feature responses from a large spatial extent, while the single-stage encoder-decoder structure encodes the high-level semantic information in the encoder path and recovers the boundary information in the decoder path. In contrast, multi-stage encoder-decoder networks have been widely used in human pose estimation and shown superior performance than their single-stage counterpart. However, few efforts have been attempted to bring this effective design to semantic segmentation. In this work, we propose a Semantic Prediction Guidance (SPG) module which learns to re-weight the local features through the guidance from pixel-wise semantic prediction. We find that by carefully re-weighting features across stages, a two-stage encoder-decoder network coupled with our proposed SPG module can significantly outperform its one-stage counterpart with similar parameters and computations. Finally, we report experimental results on the semantic segmentation benchmark Cityscapes, in which our SPGNet attains 81.1\% on the test set using only `fine' annotations. 
\end{abstract}
\section{Introduction}

\begin{figure}[t]
	\centering
	\begin{minipage}[b]{0.99\linewidth}
    \centering
    \vspace{0mm}
    \includegraphics[width=0.86\linewidth]{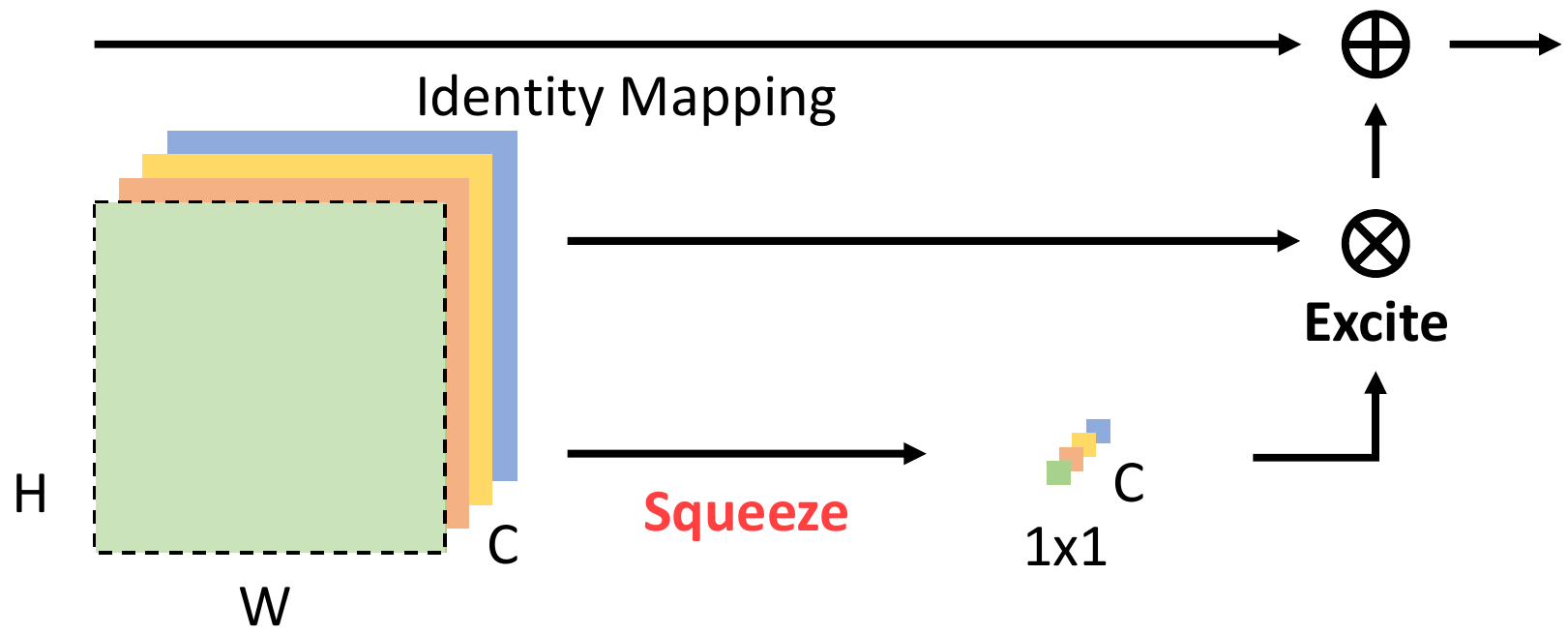}
    \\
    {\footnotesize(a) Squeeze-and-Excite~\cite{hu2018squeeze}.}
    \\
    \end{minipage}
    \begin{minipage}[b]{0.99\linewidth}
    \centering
    \vspace{0mm}
    \includegraphics[width=0.86\linewidth]{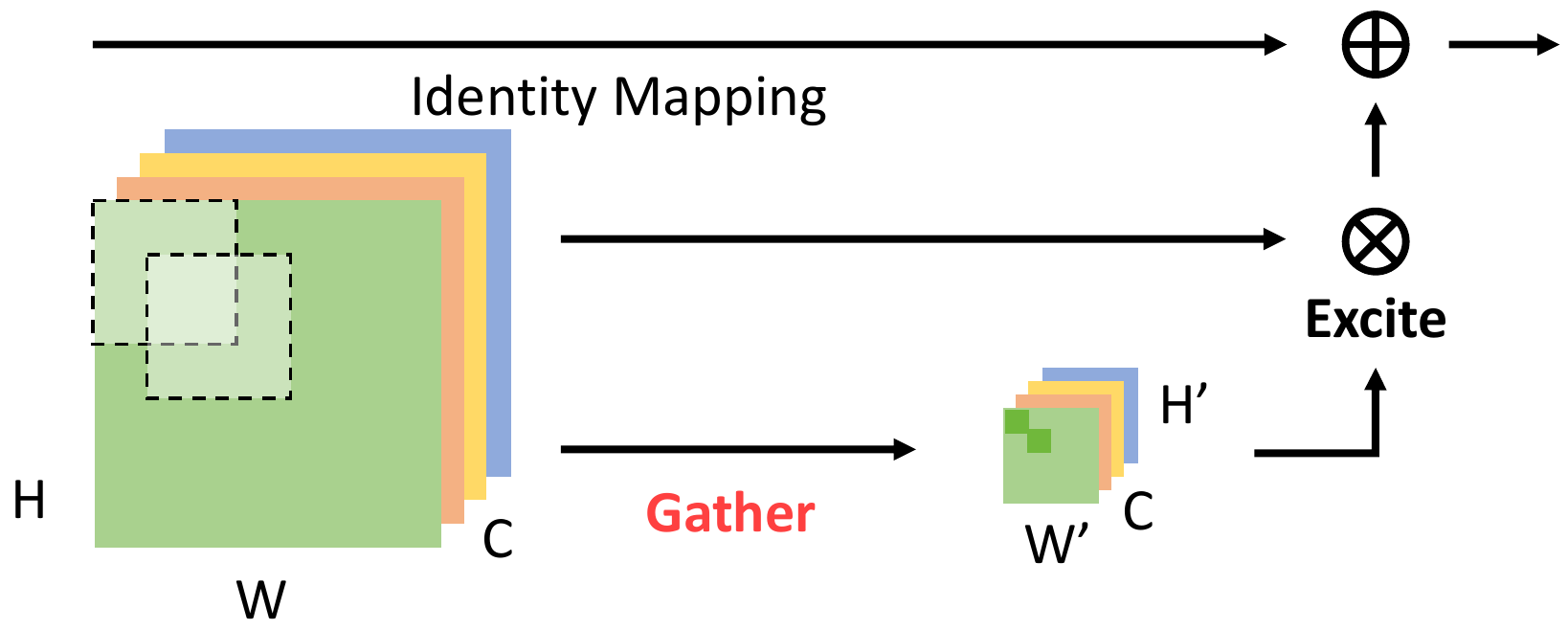}
    \\
    {\footnotesize(b) Gather-and-Excite~\cite{hu2018gather}.}
    \\
    \end{minipage}
    \begin{minipage}[b]{0.99\linewidth}
    \centering
    \vspace{0mm}
    \includegraphics[width=0.86\linewidth]{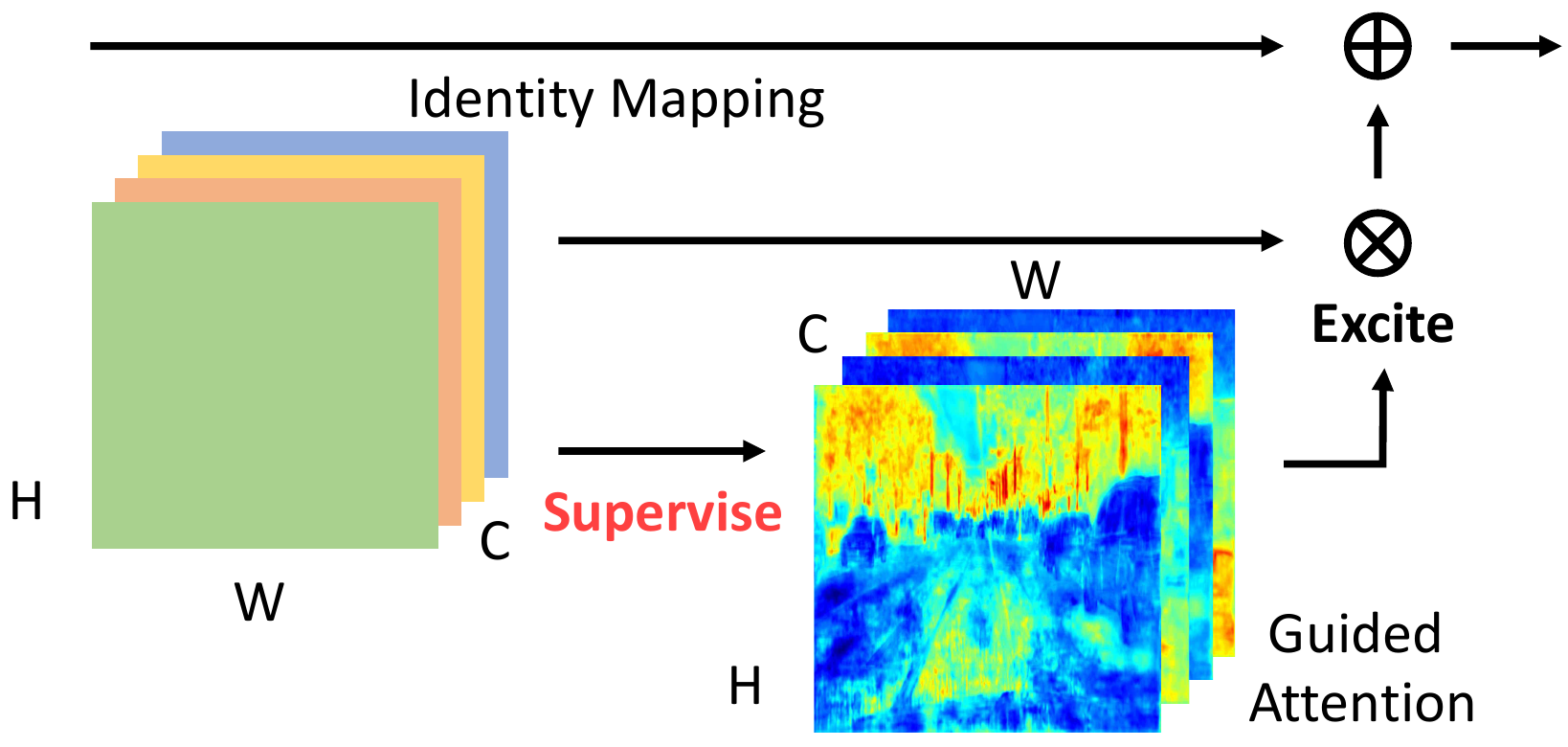}
    \\
    {\footnotesize(c) Our proposed Supervise-and-Excite.}
    \\
    \end{minipage}
	\caption{
	Three different frameworks for re-weighting local features. Dotted square areas with lighter color in (a) and (b) denote average pooling operations.
	}
	\label{fig:se}
\end{figure}

Semantic segmentation \cite{HeZC04}, as a step towards scene understanding \cite{tu2005image,ladicky2010and,yao2012describing,tighe2013finding}, is a challenging problem in computer vision. It refers to the task of assigning semantic labels, such as person and sky, to every pixel within an image.  Recently, Deep Convolutional Neural Networks (DCNNs) \cite{lecun1989backpropagation,krizhevsky2012imagenet} have significantly improved the performance of semantic segmentation systems.

In particular, DCNNs, deployed in a fully convolutional manner \cite{sermanet2013overfeat,long2015fully}, have attained remarkable results on several semantic segmentation benchmarks \cite{Everingham10,cordts2016cityscapes,zhou2017scene}. We observe two key design components shared among state-of-the-art semantic segmentation systems. First, multi-scale context module, exploiting the large spatial information, enriches the local features. Typical examples include DeepLab \cite{deeplabv12015} which adopts several parallel atrous convolutions \cite{holschneider1989real,papandreou2014untangling} with different rates and PSPNet \cite{zhao2017pyramid} which performs pooling operations at different grid scales. Recently, SENets \cite{hu2018squeeze} and GENets \cite{hu2018gather} employ the `squeeze-and-excite' (Figure~\ref{fig:se}~(a)) or more general `gather-and-excite' framework (Figure~\ref{fig:se}~(b)) and obtain remarkable results on image classification task. Motivated by this, we propose a simple yet effective attention module, called Semantic Prediction Guidance (SPG), which learns to re-weight the local feature map values via the guidance from pixel-wise semantic prediction. Unlike the `gather-and-excite' module \cite{hu2018squeeze,hu2018gather} (where context information is gathered from a large spatial extent and local features are excited accordingly), our SPG module adopts the `\textbf{supervise-and-excite}' framework (Figure~\ref{fig:se}~(c)). Specifically, we inject the semantic supervision to the feature maps followed by a simple $1 \times 1$ convolution with sigmoid activation function (\ie, `supervise' step). The resulting feature maps, called ``Guided Attention'', are used as a guidance to re-weight the other transformed feature maps correspondingly (\ie, `excite' step). We further add an `identity' mapping in the module, similar to the residual block \cite{he2016deep}. Additionally, our learned ``Guided Attention'' allows us to visually explain the ``re-weighting'' mechanism in our SPG module.

Another important design component is the encoder-decoder structure, where high-level semantic information is captured in the encoder path while the detailed low-level boundary information is recovered in the decoder path. The systems \cite{noh2015learning,ronneberger2015u,badrinarayanan2017segnet,lin2017refinenet,pohlen2017full,peng2017large,amirul2017gated,zhang2018exfuse,xiao2018unified,deeplabv3plus2018,wojna2019devil}, employing the single-stage encoder-decoder structure (\ie, the encoder-decoder structure is stacked only once), have demonstrated outstanding performance on several semantic segmentation benchmarks. On the other hand, the multi-stage encoder-decoder models \cite{wei2016convolutional,cao2017realtime,newell2017associative,newell2016stacked,yang2017learning,ke2018multi,li2019rethinking}, also known as stacked hourglass networks \cite{newell2016stacked}, refine the keypoint estimation iteratively by propagating information across stages for the task of human pose estimation. Interestingly, we observe that the multi-stage encoder-decoder structure is seldom explored in the context of semantic segmentation, except \cite{fu2019stacked,shah2018stacked}. In this work, we revisit the multi-stage encoder-decoder networks on the Cityscapes dataset \cite{cordts2016cityscapes}. We find that by carefully selecting features across stages, a two-stage encoder-decoder network coupled with our proposed SPG module can significantly outperform its one-stage counterpart with similar parameters and computations.

On Cityscapes test set \cite{cordts2016cityscapes}, our proposed SPGNet outperforms the strong baseline DenseASPP \cite{yang2018denseaspp} when only exploiting the `fine' annotations. Our overall mIoU is slightly behind the concurrent work DANet \cite{fu2018dual} but detailed class-wise mIoU reveals that our model is better than DANet in 14 out of 19 semantic classes. Furthermore, our SPGNet requires only $22.7\%$ computation of DANet \cite{fu2018dual}.

To summarize our main contributions:
\begin{itemize}
\setlength\itemsep{0em}
\item We propose a simple yet effective attention module, called SPG, which adopts a `supervise-and-excite' framework.
\item We explore multi-stage encoder-decoder networks on semantic segmentation task. Incorporating our proposed SPG module to the multi-stage encoder-decoder networks further improves the performance.
\item We demonstrate the effectiveness of our SPGNet on the challenging Cityscapes dataset. Our model outperforms the strong baseline DenseASPP \cite{yang2018denseaspp}, and is better than DANet \cite{fu2018dual} in 14 out of 19 semantic classes. Our SPGNet strikes a better accuracy/speed trade-off, requiring only $22.7\%$ computation of DANet.
\item We provide detailed ablation studies along with the visualization of our learned attention maps. We also discuss the effectiveness of employing multi-stage encoder-decoder networks on semantic segmentation.
\end{itemize}

\begin{figure*}[!t]
	\centering
	\includegraphics[width=1.0\linewidth]{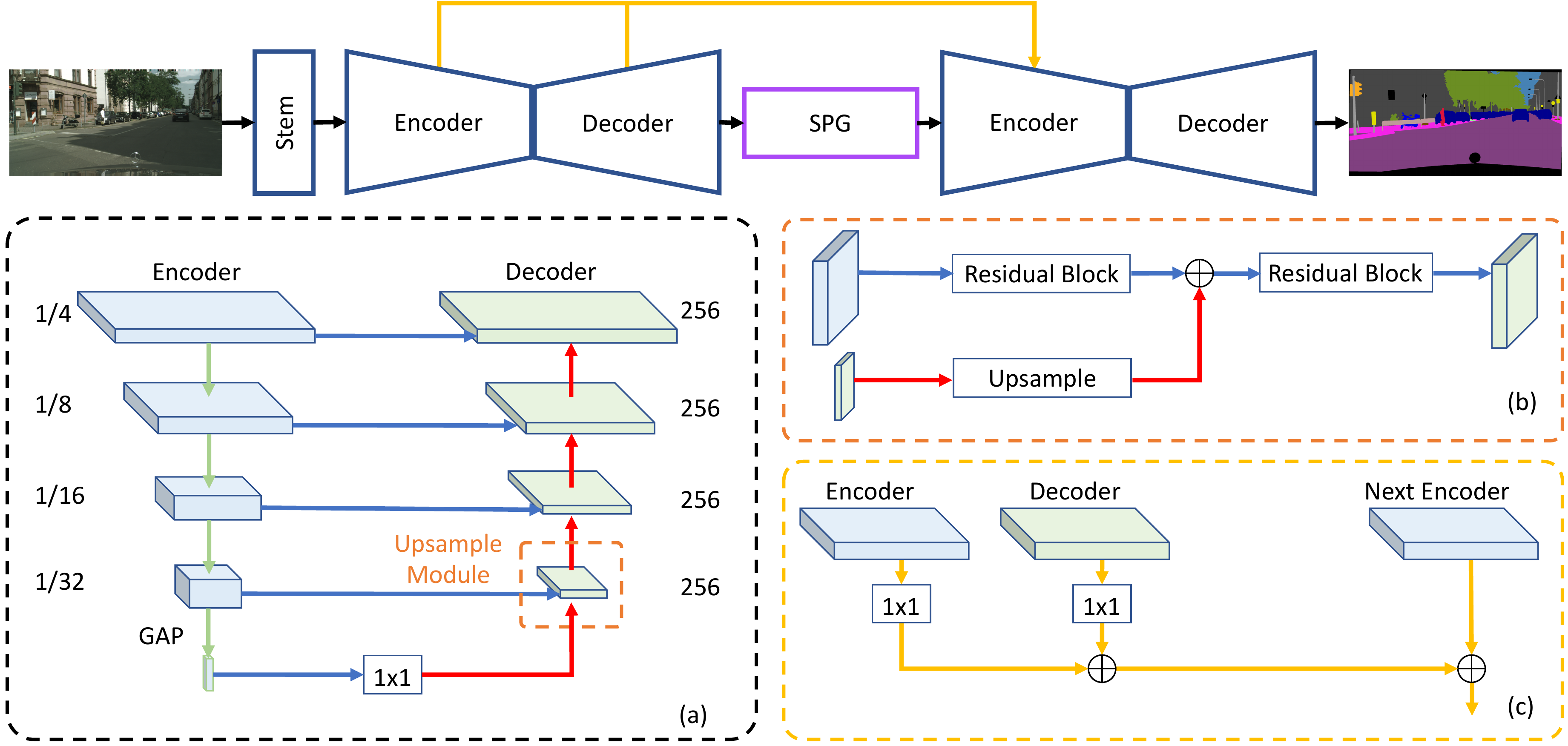}
	\caption{
	The overall structure of SPGNet. Only two stages are shown for simplicity and it can be easily generalized to more stages. (a) Our encoder-decoder design. (b) Upsample module. (c) Cross stage feature aggregation~\cite{li2019rethinking}. GAP: global average pooling \cite{liu2015parsenet}. Residual Block: same bottleneck module used in ResNet~\cite{he2016deep}. Upsample: bilinear upsampling by x2. SPG: semantic prediction guidance module.
	}
	\label{fig:overall_structure}
\end{figure*}

\section{Related Works}
\noindent{\bf Semantic Segmentation:} Most state-of-the-art semantic segmentation models are based on FCN \cite{sermanet2013overfeat,long2015fully}. The detailed object boundary information is usually missing due to the pooling or convolutions with striding operations within the network. To alleviate the problem, one could  apply the atrous convolution \cite{holschneider1989real,sermanet2013overfeat,papandreou2014untangling,deeplabv12015} to extract dense feature maps. However, it is computationally expensive to extract output feature maps that are 8 or even 4 times smaller than the input resolution using state-of-the-art network backbones \cite{krizhevsky2012imagenet,simonyan2014very,szegedy2015going,he2016deep}. On the other hand, the encoder-decoder structures \cite{noh2015learning,ronneberger2015u,badrinarayanan2017segnet,lin2017refinenet,pohlen2017full,peng2017large,amirul2017gated,fu2019stacked,zhang2018exfuse,xiao2018unified,deeplabv3plus2018,wojna2019devil,jiao2019geometry} capture the context information in the encoder path and recover high resolution features in the decoder path. Additionally, contextual information has also been explored. ParseNet \cite{liu2015parsenet} exploits the global context information, while PSPNet \cite{zhao2017pyramid} uses spatial pyramid pooling at several grid scales. DeepLab \cite{chen2018deeplabv2,chen2017deeplabv3,liu2019auto,yang2019deeperlab} uses several parallel atrous convolution with different rates in the Atrous Spatial Pyramid Pooling module, while DPC \cite{dpc2018} applies neural architecture search \cite{zoph2017neural} for the context module. Finally, our proposed Semantic Prediction Guidance (SPG) bears a similarity to the Layer Cascade method \cite{li2017not} which treats each pixel differently. Instead of classifying easy pixels in the early stages within the network, our SPG module weights each pixel according to the predictions in the first stage of our stacked network.

\noindent{\bf Multi-Stage Networks:} Multi-stage networks \cite{wei2016convolutional,cao2017realtime,newell2017associative,newell2016stacked,yang2017learning,ke2018multi,li2019rethinking,song2018spg,wang2017stagewise} have been widely used and explored in human pose estimation. Multi-stage networks aim to iteratively refine estimation. To maximally utilize the capacity of each stage, CPM \cite{wei2016convolutional} and Stacked Hourglass \cite{newell2016stacked} propagate not only features to the next stage, but also remap predicted heatmaps into feature space by a 1x1 convolution and concatenate with feature maps. MSPN \cite{li2019rethinking} further optimizes feature flow across stages by propagating intermediate features of encoder-decoder of previous stage to the next stage. MSPN \cite{li2019rethinking} demonstrates superior performance over single stage counterpart with similar parameters and computations. On the other hand, Stacked Deconvolutional Network \cite{fu2019stacked} uses multiple deconvolution networks for semantic segmentation. However, it only passes features across stages and neglects predictions of every stage. Additionally, Zhou \etal \cite{zhou2017scene} propose a cascade segmentation module. In this work, we find predictions can be served as a special attention to propagate useful features across stages. 

\noindent{\bf Attention Module:} Attention mechanism has been widely used recently in multiple computer vision tasks. Chen \etal \cite{chen2015attention} learn an attention module to merge multi-scale features. Kong and Fowlkes \cite{kong2018recurrent} propose a gating module that adaptively selects
features pooled with different field sizes. Recently, the self-attention module \cite{vaswani2017attention} has been explored by several works \cite{hu2017relation,wang2018non,fu2018dual,chen2018a2,huang2019ccnet} for computer vision tasks. In contrast, our proposed SPG module is more similar to the other works that employ the `squeeze-and-excite' or `gather-and-excite' framework. In particular, Squeeze-and-Excitation Networks (SENets) \cite{hu2018squeeze} squeeze the features across spatial dimensions to aggregate the information for re-weighting feature channels. Hu \etal \cite{hu2018gather} generalize SENets with `gather-and-excite' operations where long-range spatial information is gathered to re-weight (or `excite') the local features. Motivated by this, our proposed SPG module employs the `supervise-and-excite' framework, where our local features are guided by the semantic supervision. Additionally, EncNet \cite{zhang2018context} also adds supervision to their global feature. However, our supervision is pixel-level instead of image-level.

\section{Methods}

\subsection{Overall Architecture}
Figure~\ref{fig:overall_structure} shows our proposed SPGNet, which consists of multiple stages and each stage is based on an encoder-decoder architecture: encoder produces dense feature maps at multiple scales and also an image-level feature vector using global average pooling (GAP). Decoder starts with this feature vector and gradually recovers spatial resolution by combining corresponding encoder feature map using an upsample module, described in Sec~\ref{sec:upm}. 

Our SPGNet stacks multiple stages, where earlier decoder output is fed into a semantic prediction guidance (SPG) module (detailed in Sec~\ref{sec:spg}) to generate input feature for the next stage. In addition, we employ Cross Stage Feature Aggregation~\cite{li2019rethinking} to enhance latter stage encoders by taking advantage of earlier stage encoder / decoder features, as shown in Figure~\ref{fig:overall_structure}(c). The decoder output in the final stage is bilinearly upsampled to input image resolution, generating per-pixel semantic prediction results.  

The multi-stage design of SPGNet is inspired by Stacked Hourglass~\cite{newell2016stacked} for human pose estimation. Our method differs from Stacked Hourglass in 1) we carefully design the encoder-decoder architecture in each stage instead of using a symmetric hourglass network, and 2) latter stage input is generated from SPG module rather than simply passing the features combined with predictions from previous stage.

\subsection{Encoder / Decoder Design} \label{sec:edd}
Hourglass networks~\cite{newell2016stacked} assign equal computation to both encoder and decoder, making it unavailable to use pre-trained weights on ImageNet~\cite{deng2009imagenet}. In contrast, Feature Pyramid Networks (FPN)~\cite{lin2017feature} use well-designed classification networks for encoder and design a simple decoder consisting of only nearest-neighbor interpolations to upsample decoder feature maps. Our encoder-decoder design principles follow FPN (\eg, all the feature maps in the decoder contain 256 channels), but we employ two more components to make it more efficient and effective. First, we incorporate a global average pooling \cite{liu2015parsenet} after the output of encoder to generate the $1\times1$ image-level features followed by another $1\times1$ convolution to transform its feature channels to 256. Second, instead of using a single nearest-neighbor interpolation, we design an efficient upsample module, as described in the next section.

\subsection{Upsample module} \label{sec:upm}
As illustrated in Figure~\ref{fig:overall_structure}(a, b), our decoder adopts upsample module to recover feature map resolution step-by-step. Specifically, each module in the decoder takes two input feature maps, one from encoder and one from previous layer output. The input from encoder is first transformed by a residual block to reduce the dimension to output channel of the decoder. Then, the input from previous layer output is bilinear upsampled and added to the transformed encoder output. Instead of passing this merged feature directly to next upsample module, we further add another residual block to better fuse features from two different sources.

\subsection{Semantic Prediction Guidance} \label{sec:spg}

\begin{figure}[!t]
	\centering
	\includegraphics[width=1.0\linewidth]{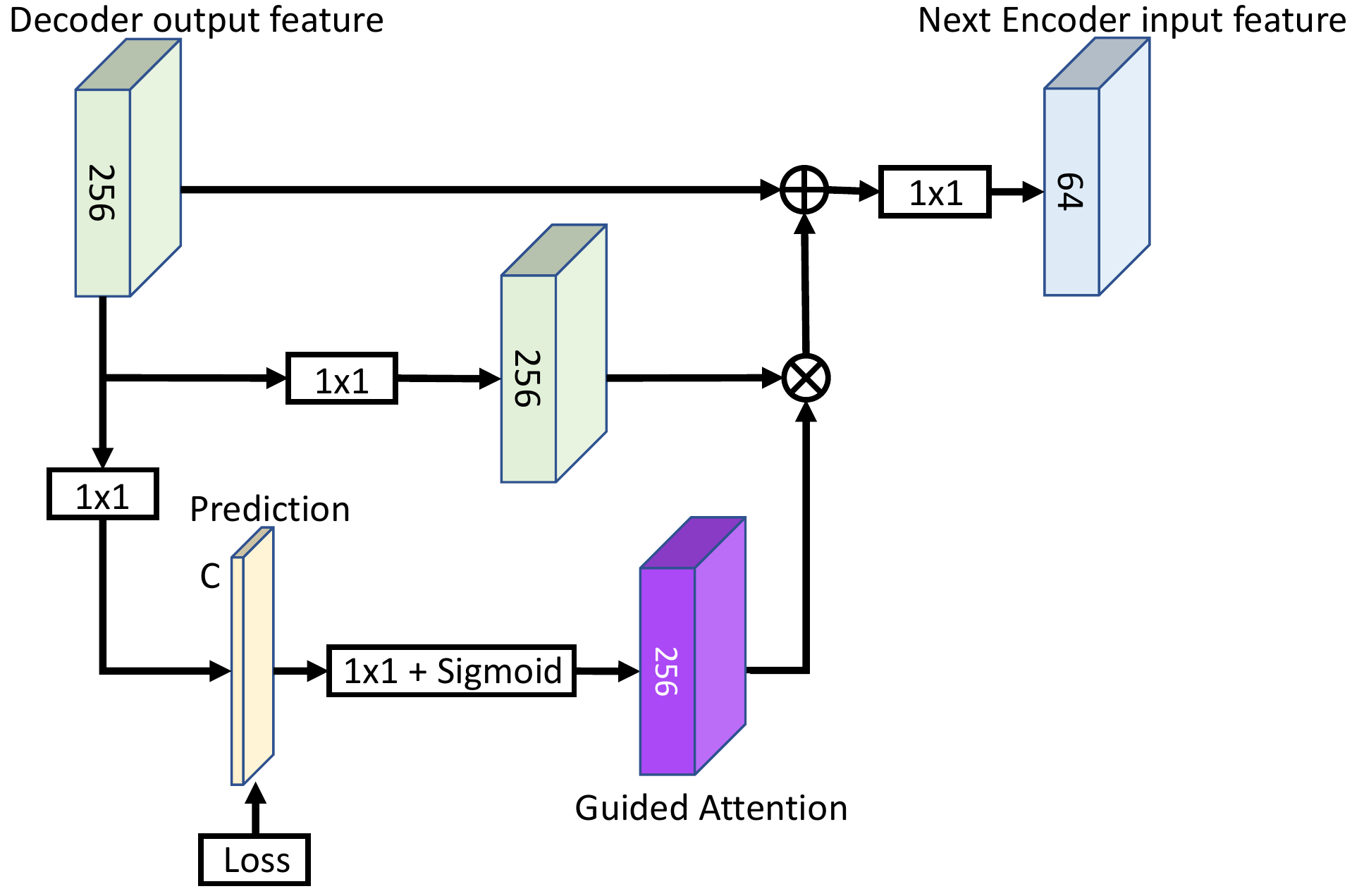}
	\caption{
	Our Semantic Prediction Guidance (SPG) module.
	}
	\label{fig:spg}
\end{figure}

Using contextual information to re-weight feature channels~\cite{hu2018squeeze,hu2018gather} has brought significant improvements to image classification task. This process usually includes a `gather' step which collects information over a large spatial region. In contrast, in our multi-stage encoder-decoder network, the output features generated from each stage already contain information from multiple scales. This inspires us to design a simple yet effective SPG module (Figure~\ref{fig:spg}) which treats the features from earlier stage as `gathered' information. Specifically, the previous stage decoder output feature $x_d \in \mathbb{R}^{H\times W \times D}$ is first fed into a $1\times 1$ convolution to produce per-class logits $x_l \in \mathbb{R}^{H\times W \times C}$, where $H$ and $W$ are decoder output height and width, $D$ is the number of channels used in the decoder and $C$ is the number of semantic classes in the dataset. We then produce a per-pixel, per-channel Guided Attention mask $m \in \mathbb{R}^{H\times W \times D} $ from $x_l$ via a simple $1\times 1$ convolution followed by sigmoid activation. This Guided Attention will be element-wise multiplied to a transformed decoder feature map, generated from $1\times 1$ convolution on top of $x_d$, resulting in an attention-augmented feature map. Similar to residual block, this feature map is added back to decoder output feature $x_d$, followed by another $1\times 1$ convolution to produce input feature to the next stage encoder. During training, we minimize the loss of last-stage semantic prediction and per-class logits $x_l$ in all previous stages.

Our proposed SPG module differs from SENets~\cite{hu2018squeeze} and GENets~\cite{hu2018gather} on using supervised semantic predictions to guide the `excite' step. We further verified having explicit supervision improves model performance.
The benefit of our proposed SPG module are twofold: the `gather' step is implicitly folded into the encoder-decoder architecture, which allows SPG module to be computationally efficient (about $1\%$ increase in FLOPs) and have a small memory footprint ($2.3\%$ higher peak memory usage). Meanwhile, using semantic prediction makes SPG module more explainable. See Section~\ref{sec:vis} for visualization.

\section{Experiments}

\subsection{Dataset}
We perform experiments on the Cityscapes dataset \cite{cordts2016cityscapes}, which contains 19 classes. There are 5,000 images with high quality annotation (called ``fine''), divided into 2,975/500/1,525 images for training, validation and testing. We only use the ``fine'' annotation in this paper.

\subsection{Implementation Details}

\noindent{\bf Networks.} %
We employ ResNet \cite{he2016deep} in the encoder module. The ``Stem'' in Figure~\ref{fig:overall_structure} consists of a $7\times7$ convolution with stride $=2$ followed by a $3\times3$ max pooling with stride $=2$. We replace BatchNorm layers with synchronized Inplace-ABN \cite{rota2018place}, and adopt bilinear interpolation in all the upsampling operations.

\noindent{\bf Training settings.} We use mini-batch SGD momentum optimizer with batch size 8, initial learning rate 0.01, momentum 0.9 and weight decay 0.0001. Following prior works \cite{liu2015parsenet}, we use the ``poly'' learning rate schedule where the learning rate is scaled by $(1 - \frac{\text{iter}}{\text{iter}_{\text{max}}})^{0.9}$. For data augmentation, we employ random scale between [0.5, 2.0] with a step size of 0.25, random flip and random crop. We train the model for $80,000$ iterations on ``train''set for ablation study. To evaluate our model on the ``test'' set, we train the model on the concatenation of ``train'' and ``val'' set.

\subsection{Comparison with State-of-the-Arts}
\begin{table}[!t]\setlength{\tabcolsep}{6pt}
\centering
\begin{threeparttable}
\scalebox{0.76}{
\begin{tabular}{l c c c c}
\toprule[0.2em]
Method  & Backbone & mIoU ($\%$) & \#Params & \#FLOPs\tnote{\textdagger} \\
\toprule[0.2em]
RefineNet~\cite{lin2017refinenet} &   ResNet-$101$  &  $73.6$ & - & -\\ 
DUC-HDC~\cite{wang2018understanding} &  ResNet-$101$ & $77.6$ & $65.0$M & $2234.3$B\\ 
SAC~\cite{zhang2017scale}  &   ResNet-$101$  &  $78.1$ & - & - \\
DepthSeg~\cite{kong2018recurrent} &   ResNet-$101$  &  $78.2$ & -  & - \\ 
PSPNet~\cite{zhao2017pyramid}  &   ResNet-$101$  &  $78.4$ & $65.7$M & $2117.3$B \\ 
BiSeNet~\cite{yu2018bisenet} &   ResNet-$101$  &  $78.9$ & $51.6$M & $429.5$B\\ 
DFN~\cite{yu2018learning} &   ResNet-$101$  &  $79.3$ & $112.0$M & $2239.6$B\\ 
PSANet~\cite{zhao2018psanet} &   ResNet-$101$  &  $80.1$ & -  & - \\ 
DenseASPP~\cite{yang2018denseaspp}  &   DenseNet-$161$  &  $80.6$ & $35.4$M & $1240.1$B\\ 
DANet~\cite{fu2018dual} & ResNet-$101$ & \underline{$81.5$} & $66.5$M & $2878.9$B\\ 
\midrule
SPGNet (Ours) &   $2\times$ ResNet-$50$ & $\bf{81.1}$ & $59.8$M & $654.8$B\\
\bottomrule[0.1em]
\end{tabular}
}
\begin{tablenotes}
\item[\textdagger] {\scriptsize \#FLOPs take all matrix multiplication into account.}
\end{tablenotes}
\caption{Comparison to state-of-the-art on Cityscapes {\it test} set.}
\label{table:cityscapes_sota}
\end{threeparttable}
\end{table}

\begin{table*}[!t]
\renewcommand\arraystretch{1.2}
\setlength{\tabcolsep}{4pt}
    \begin{center}
    \begin{adjustbox}{max width=\textwidth}
        \begin{threeparttable}
        \begin{tabular}{ l | c |c c c c c c c c c c c c c c c c c c c  c}
            \toprule[0.2em]
       Methods &  \rotatebox{90}{Mean IoU} &  \rotatebox{90}{road} &  \rotatebox{90}{sidewalk} &  \rotatebox{90}{building} & \rotatebox{90}{ wall} &  \rotatebox{90}{fence} &  \rotatebox{90}{pole} & \rotatebox{90}{traffic light} &  \rotatebox{90}{traffic sign}&  \rotatebox{90}{vegetation} &  \rotatebox{90}{terrain} &  \rotatebox{90}{sky} & \rotatebox{90}{person} &  \rotatebox{90}{rider} & \rotatebox{90}{car} &  \rotatebox{90}{truck}& \rotatebox{90}{ bus}& \rotatebox{90}{ train}& \rotatebox{90}{ motorcycle}&  \rotatebox{90}{bicycle}\\
      \toprule[0.2em]
       DenseASPP~\cite{yang2018denseaspp} & 80.6 & 98.7 & 87.1 & 93.4 & \textbf{60.7} & 62.7 & 65.6 & 74.6 & 78.5 & 93.6 & 72.5 & 95.4 & 86.2 & 71.9 & 96.0 & \textbf{78.0} & \textbf{90.3} & 80.7 & 69.7 & 76.8\\         
       DANet \cite{fu2018dual} & \underline{81.5} & 98.6 & 86.1 & 93.5 & 56.1 & \textbf{63.3} & 69.7 & 77.3 & 81.3 & 93.9 & 72.9 & 95.7 & 87.3 & 72.9 & 96.2 & 76.8 & 89.4 & \textbf{86.5} & \textbf{72.2} & 78.2\\ 
\midrule
       SPGNet (ours) & \textbf{81.1} & \textbf{98.8} & \textbf{87.6} & \textbf{93.8} & 56.5 & 61.9 & \textbf{71.9} & \textbf{80.0} & \textbf{82.1} & \textbf{94.1} & \textbf{73.5} & \textbf{96.1} & \textbf{88.7 }& \textbf{74.9} & \textbf{96.5} & 67.3 & 84.8 & 81.8 & 71.1 & \textbf{79.4}\\
        \bottomrule[1pt]
        \end{tabular}
        \end{threeparttable}
    \end{adjustbox}
    \end{center}
\caption{Per-class results on Cityscapes test set. SPGNet outperforms existing top approaches in 13 out of 19 classes. }
\label{table:cityscapes_sota_detailed}
\end{table*}
In Table~\ref{table:cityscapes_sota}, we report our Cityscape ``test'' set result. We only use ``fine'' annotations and thus compare with the other state-of-art models that adopt the same setting in the table. Similar to other models, we use the multi-scale inputs (scales = $\{0.75, 1.0, 1.25, 1.5, 1.75, 2.0 \}$) during inference. We also report the model parameters and computation FLOPs (\wrt, a single $1024\times2048$ input size).

Our best SPGNet model variant employs a 2-stage encoder-decoder structures with ResNet-50 as encoder backbone and decoder channels $=256$. Our model outperforms most top-performing approaches on Cityscapes with much less computation. Notably, most state-of-the-art methods are mainly based on systems using atrous convolutions to preserve feature maps resolution, which however requires a large amount of computation (as indicated by \#FLOPs in Table~\ref{table:cityscapes_sota}). On the contrary, our proposed SPGNet, built on top of an efficient encoder-decoder structure, strikes a better trade-off between accuracy and speed.

To be concrete, the computation of our SPGNet is almost half of DenseASPP, the previous published state-of-the-art model using only fine annotations, but our performance is $0.5\%$ mIoU better. We also compare our SPGNet with another concurrent work DANet~\cite{fu2018dual}. Our computation is around $22.7\%$ of DANet with only $0.4$ mIoU degradation.

We further compare per-class results with the top-2 performing approaches in Table~\ref{table:cityscapes_sota_detailed}. Surprisingly, our SPGNet outperforms DenseASPP in 15 out of 19 classes and DANet in 14 out of 19 classes. The main degradation of our overall mIoU comes from the ``truck'' class which is $10.7$ IoU worse than DenseASPP and $9.5$ IoU worse than DANet. We think it is because there are only few ``truck'' annotations in Cityscapes and our SPGNet requires supervision for learning the guided attention. 

\subsection{Ablation Studies}
Here, we provide ablation studies on Cityscapes val set.

\noindent{\bf Effect of SPG module.}
\begin{table}[!t]\setlength{\tabcolsep}{6pt}
\centering
\scalebox{0.78}{
\begin{tabular}{c  c  c  c  c  c c}
 \toprule[0.2em]
 \#Stage    &  SPG    & Id. & Sup. & mIoU ($\%$) & \#Params & \#FLOPs \\
 \toprule[0.2em]
 $1$      &    -       & - & - & $74.48$     & $11.7$M  & $107.6$B \\
 \midrule
 $2$      &   \xmark       & - & - & $76.31$     & $23.9$M  & $215.5$B \\ 
 \midrule
 $2$      &   \cmark (sum) & \cmark & \cmark & $76.96$     & $23.9$M  & $218.0$B \\ 
 $2$      &   \cmark (softmax) & \cmark & \cmark& $77.17$     & $23.9$M  & $218.0$B \\ 
 $2$      &   \cmark (sigmoid) & \cmark & \cmark& $\textbf{77.67}$     & $23.9$M  & $218.0$B \\ 
 \midrule
 $2$      &   \cmark (sigmoid) & \xmark & \cmark & $77.24$     & $23.9$M  & $218.0$B \\
 $2$      &   \cmark (sigmoid) & \cmark & \xmark & $77.12$     & $23.9$M  & $218.0$B \\ 
 \bottomrule[0.1em]
\end{tabular}
}
\caption{Cityscapes ablation studies of proposed SPG on validation set. All models use ResNet-18 in encoder. {\bf Id.:} Adding the identity mapping path in SPG. {\bf Sup.:} Adding the semantic supervision in SPG. Employing sigmoid activation, identity mapping path, and supervision in our SPG with 2 stages attains the best performance.}
\label{table:cityscapes_ablation}
\end{table}
We perform ablation studies on the SPG design in Table~\ref{table:cityscapes_ablation}. The baseline is a simple 2-stage encoder-decoder network by directly passing the 1st stage decoder features to the 2nd stage encoder. This baseline model uses the Cross-Stage Feature Aggregation (CSFA)~\cite{li2019rethinking} which is slightly better than the case without CSFA by 0.18\%. We first verify whether passing semantic prediction together with the decoder features to next stage is helpful. We transform the predictions from the 1st stage decoder output by applying a $1\times1$ convolution. The sum of the transformed predictions and the 1st stage decoder output is passed to the next stage (denoted as SPG (sum)). It achieves $76.96\%$ mIoU which is $0.65\%$ mIoU better than the baseline. Additionally, our proposed SPG module uses the transformed semantic predictions to `excite' the decoder features. We explore two ways for excitation: one is by applying softmax on the spatial dimension $\text{H}\times\text{W}$ (SPG (softmax)) and the other is using sigmoid (SPG(sigmoid)). The SPG (softmax) scheme improves the baseline by $0.86\%$ mIoU while the SPG (sigmoid) scheme achieves the best mIoU of $77.67\%$ ($1.36\%$ mIoU better than the baseline). Comparing results of SPG (sigmoid) scheme ($77.67\%$ mIoU) with SPG (sum) scheme ($76.96\%$ mIoU), it shows the importance of using `Excite' to re-weight features. Finally, we investigate the effect of adding the identity mapping path and the supervision in SPG module. Dropping the identity mapping path in Figure~\ref{fig:spg} degrades the performance from $77.67\%$ to $77.24\%$, while removing the supervision on learning the guided attention decreases the performance to $77.12\%$ in which our SPG module degenerates to a special case of `gather-and-excite' (where the features are `gathered` from the 1st-stage decoder output).

\noindent{\bf SPG module \emph{vs} SE/GE module.}
\begin{table}[!t]\setlength{\tabcolsep}{18pt}
\centering
\scalebox{0.9}{
\begin{tabular}{c c c}
\toprule[0.2em]
 Module & Supervise    & mIoU ($\%$) \\
\toprule[0.2em]
 SE~\cite{hu2018squeeze} & \xmark & $77.09$    \\
 GE~\cite{hu2018gather} & \xmark & $77.22$    \\
 \midrule
 SPG (Ours) & \cmark & $\textbf{77.67}$    \\
\bottomrule[0.1em]
\end{tabular}
}
\caption{Cityscapes val ablation studies on Supervise-and-Excite. All models use ResNet-18 in encoder. Our proposed Supervise-and-Excite has advantage over Squeeze/Gather-and-Excite.}
\label{table:cityscapes_se}
\end{table}
To demonstrate the gain of SPG module comes from supervision, we compare SPG module with its unsupervised counterpart, \ie SE \cite{hu2018squeeze} and GE \cite{hu2018gather} modules. Using SE and GE modules achieves 77.09 mIoU and 77.22 mIoU respectively, both results are better than the baseline 76.31 mIoU and using GE is slightly better than SE which is consistent with the findings in \cite{hu2018gather}. However, they are still worse than using our proposed supervise-and-excite (\ie SGP with 77.67 mIoU). The additional gain mainly comes from adding supervision in supervise-and-excite.

\noindent{\bf More stages.}
\begin{table}[!t]\setlength{\tabcolsep}{12pt}
\centering
\scalebox{1}{
\begin{tabular}{c c c c}
\toprule[0.2em]
\#Stage    & mIoU ($\%$) & \#Params & \#FLOPs \\
\toprule[0.2em]
 $1$      & $74.48$     & $11.7$M  & $107.6$B \\
 $2$      & $\textbf{77.67}$     & $23.9$M  & $218.0$B \\ 
 $3$      & $77.66$     & $36.2$M  & $328.5$B \\ 
\bottomrule[0.1em]
\end{tabular}
}
\caption{Cityscapes ablation studies on validation set. All models use ResNet-18 in encoder. SPG with 2 stages is the optimal choice. 
}
\label{table:cityscapes_stacks}
\end{table}
We experiment the effect of using more stages and the results are shown in Table~\ref{table:cityscapes_stacks}. Similar to the situation in pose estimation that performance gets saturated as the number of stages increases. But in our case the performance saturates very quickly and achieves optimal with 2 stages. It is possible that by carefully balancing the loss weights among stages the performance might be better for models with more than 2 stages. However, for simplicity, we focus on models with only 2 stages in this paper. 

\noindent{\bf Effect of encoder combination.}
\begin{table}[!t]
\centering
\scalebox{0.9}{
\begin{tabular}{c c c c}
\toprule[0.2em]
 Encoder combination    & mIoU ($\%$) & \#Params & \#FLOPs \\
\toprule[0.2em]
 ResNet-$18$ + ResNet-$18$      &  $77.67$     & $23.9$M  & $218.0$B \\ 
 ResNet-$18$ + ResNet-$50$      &  $78.34$     & $38.4$M  & $336.0$B \\ 
 ResNet-$50$ + ResNet-$18$      &  $78.83$     & $38.0$M  & $329.9$B \\ 
 ResNet-$50$ + ResNet-$50$      &  $79.81$     & $55.6$M  & $467.6$B \\ 
\bottomrule[0.1em]
\end{tabular}
}
\caption{Cityscapes val ablation studies of encoder combination (ResNet-18 and ResNet-50). In our {\it two-stage} SPGNet, it is effective to employ a deeper backbone in the 1st encoder module.}
\label{table:cityscapes_quality}
\end{table}
Our {\it two-stage} SPGNet could potentially employ two different backbones in each encoder module. As shown in Table~\ref{table:cityscapes_quality}, although employing ResNet-$18$+ResNet-$50$ (\ie., ResNet-18 in the 1st encoder and ResNet-50 in the 2nd encoder) and ResNet-$50$+ResNet-$18$ have similar parameters and computation, using deeper model in the first stage outperforms the other one. We think it is crucial to ``encode'' the features in the early stage with a stronger backbone. Adopting R-$50$+R-$50$ achieves the best performance. For simplicity, we only adopt the same network backbones in all the encoder modules in this paper.

\noindent{\bf Encoder depth.}
\begin{table}[!t]\setlength{\tabcolsep}{5pt}
\centering
\scalebox{0.82}{
\begin{tabular}{c c c c c c}
\toprule[0.2em]
 Backbone & \#Stage & Channel &  mIoU ($\%$) & \#Params & \#FLOPs \\
\toprule[0.2em]
 ResNet-18 & $1$   & 128   & $74.48$     & $11.7$M  & $107.6$B \\
 ResNet-50 & $1$   & 128   & $77.80$     & $24.7$M  & $212.9$B \\
 ResNet-101 & $1$  & 128    & $78.72$     & $43.7$M  & $371.7$B \\
 ResNet-152 & $1$  & 128    & $78.33$     & $59.4$M  & $530.1$B \\
\midrule
 ResNet-18 & $2$   & 128   & $77.67$     & $23.9$M  & $218.0$B \\
 ResNet-50 & $2$   & 128   & $79.81$     & $55.6$M  & $467.6$B \\ 
 ResNet-101 & $2$  & 128    & $\textbf{80.04}$     & $93.5$M  & $785.3$B \\ 
\bottomrule[0.1em]
\end{tabular}
}
\caption{Cityscapes val ablation studies on encoder depth. Using deeper encoder in general has better performance.}
\label{table:cityscapes_encoder_depth}
\end{table}
In Table~\ref{table:cityscapes_encoder_depth}, we study the effect of adopting different backbones in the encoder module(s). We observe that using deeper encoder improves the result and using ResNet-50 in a {\it 2-stage} SPGNet achieves a good trade-off between \#Params, \#FLOPs and performance.

\noindent{\bf Hard example mining.}
\begin{table}[!t]
\centering
\scalebox{0.95}{
\begin{tabular}{c c c c c}
\toprule[0.2em]
 Backbone & \#Stage & Channel & OHEM &  mIoU ($\%$) \\
\toprule[0.2em]
 ResNet-50 & $2$   & 128   & \xmark & $79.81$    \\ 
 ResNet-50 & $2$  & 128    & \cmark  & $\textbf{80.10}$      \\ 
\midrule
 ResNet-101 & $2$   & 128   & \xmark & $80.04$    \\ 
 ResNet-101 & $2$  & 128    & \cmark  & $\textbf{80.85}$     \\ 
\bottomrule[0.1em]
\end{tabular}
}
\caption{Cityscapes val ablation studies on On-line Hard Example Mining (OHEM). SPG benefits from OHEM.}
\label{table:cityscapes_ohem}
\end{table}
We study the effects of on-line hard example (or pixel) mining (OHEM) \cite{wu2016bridging,bulo2017loss,yang2019deeperlab} in Table~\ref{table:cityscapes_ohem}. We apply OHEM to all stages (\ie, the decoder output in each stage) in our SPGNet. As shown in the table, using OHEM consistently improves the performance.

\noindent{\bf Decoder channels.}
\begin{table}[!t]\setlength{\tabcolsep}{7pt}
\centering
\scalebox{0.9}{
\begin{tabular}{c c c c c}
\toprule[0.2em]
 Backbone & Channel & mIoU ($\%$) & \#Params & \#FLOPs \\
\toprule[0.2em]
 ResNet-50 &   128    & $80.10$   & $55.6$M & $467.6$B   \\ 
 ResNet-50 &   256   & $\textbf{80.91}$  & $59.8$M & $654.8$B  \\ 
\midrule
 ResNet-101 &   128    & $80.85$   & $93.5$M & $785.3$B   \\ 
 ResNet-101 &   256   & $80.42$  & $97.8$M & $972.4$B  \\ 
\bottomrule[0.1em]
\end{tabular}
}
\caption{Cityscapes val ablation studies on decoder channel. All models are \#Stage=2 and use OHEM in training.}
\label{table:cityscapes_decoder_channel}
\end{table}
We experiment on the effect of decoder channels in Table~\ref{table:cityscapes_decoder_channel}. Employing  ResNet-50 as the encoder backbone and decoder channels $=256$ achieves the best validation mIoU.

\noindent{\bf Flip and multi-scale test.}
We further add flip and multi-scale test to the best model (ResNet-50 with 2 stages, in Table~\ref{table:cityscapes_decoder_channel}). By adding scales = $\{0.75, 1.0, 1.25, 1.5, 1.75, 2.0 \}$, the performance further improves from 80.91 to 81.86.

\begin{figure}[t]
	\centering
	\includegraphics[width=1.0\linewidth]{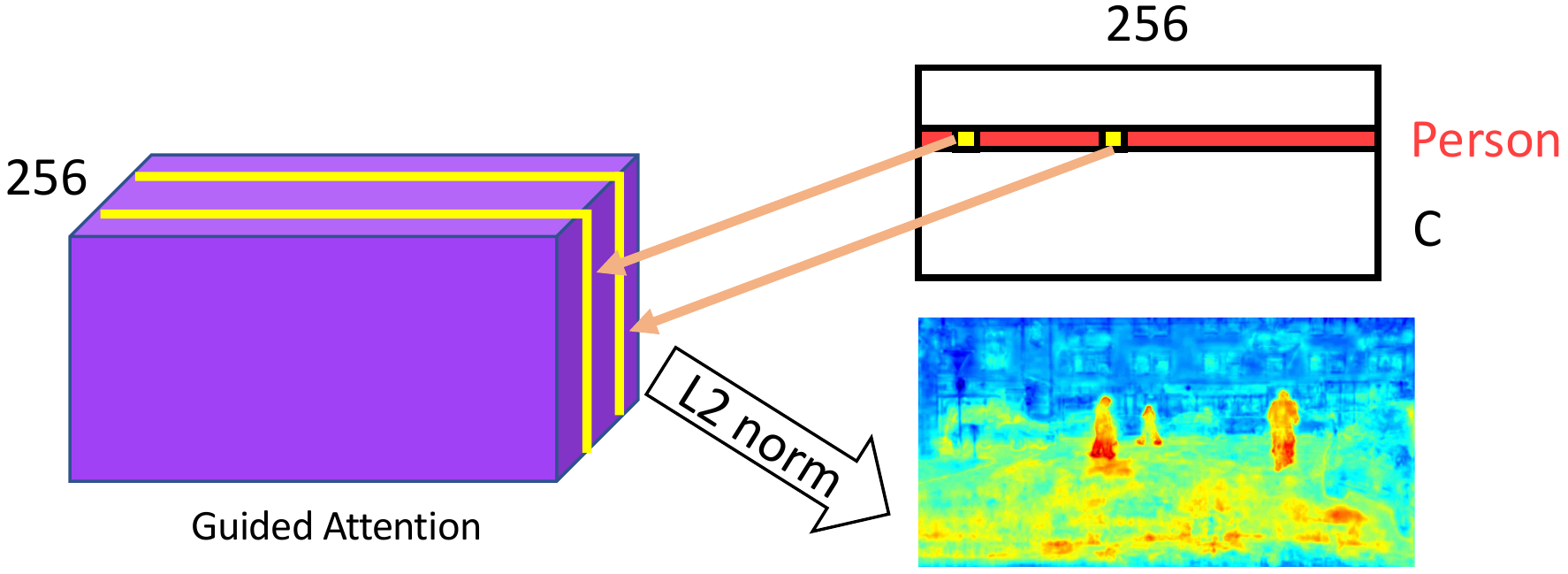}
	\caption{
	Method to visualize guided attention.
	}
	\label{fig:vis_method}

\end{figure}

\begin{figure*}[!t]
\begin{center}
\bgroup 
 \def\arraystretch{0.2} 
 \setlength\tabcolsep{0.2pt}
\begin{tabular}{cccccc}
\includegraphics[width=0.16666666666667\linewidth]{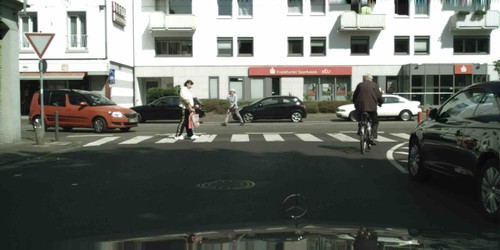} &
\includegraphics[width=0.16666666666667\linewidth]{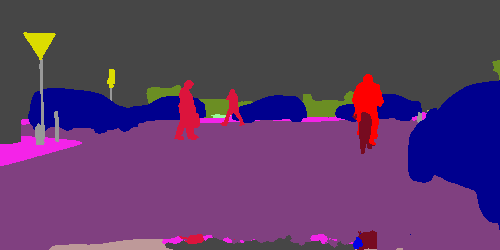} &
\includegraphics[width=0.16666666666667\linewidth]{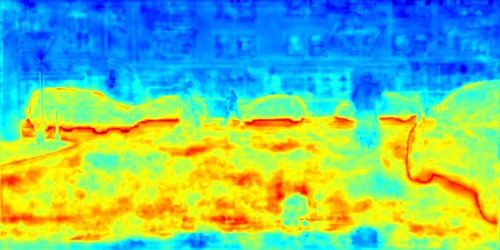} &
\includegraphics[width=0.16666666666667\linewidth]{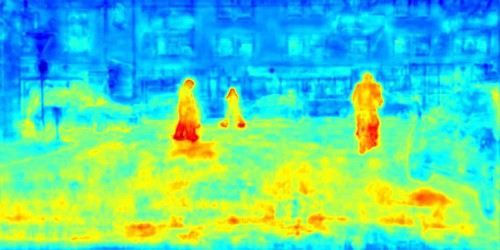} &
\includegraphics[width=0.16666666666667\linewidth]{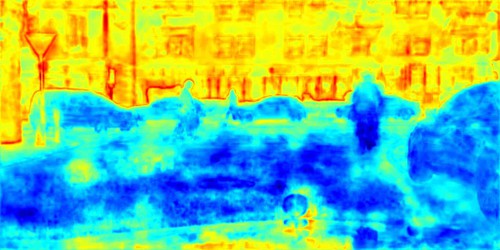} &
\includegraphics[width=0.16666666666667\linewidth]{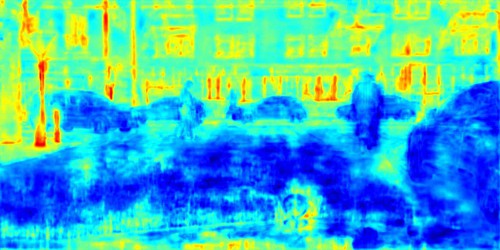} \\
\includegraphics[width=0.16666666666667\linewidth]{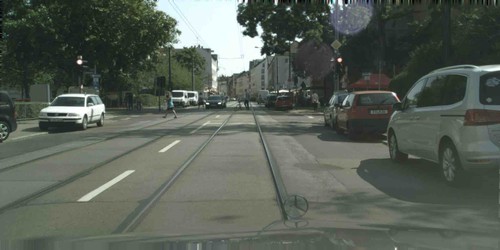} &
\includegraphics[width=0.16666666666667\linewidth]{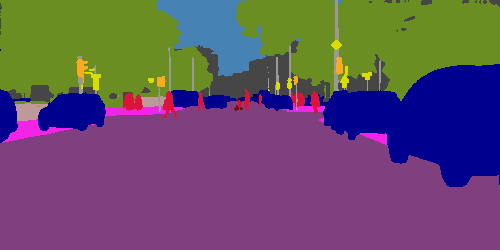} &
\includegraphics[width=0.16666666666667\linewidth]{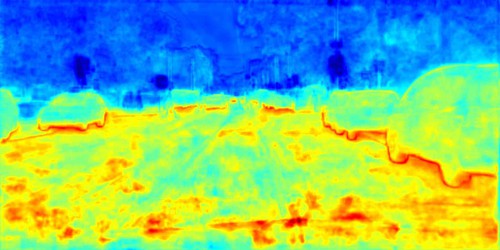} &
\includegraphics[width=0.16666666666667\linewidth]{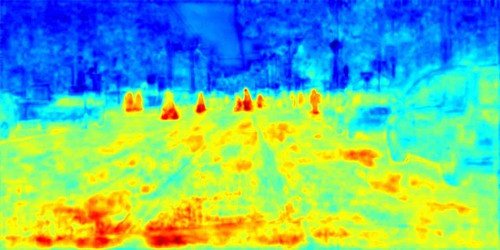} &
\includegraphics[width=0.16666666666667\linewidth]{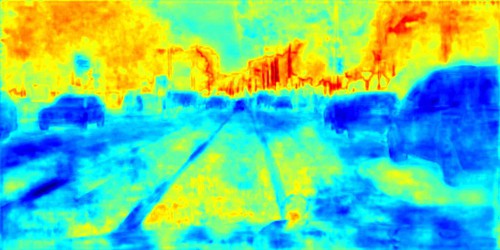} &
\includegraphics[width=0.16666666666667\linewidth]{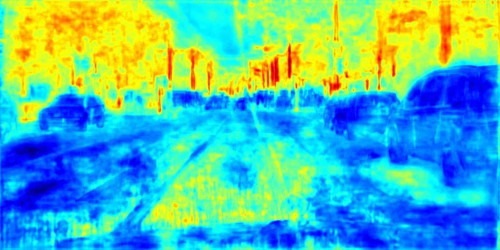} \\
\includegraphics[width=0.16666666666667\linewidth]{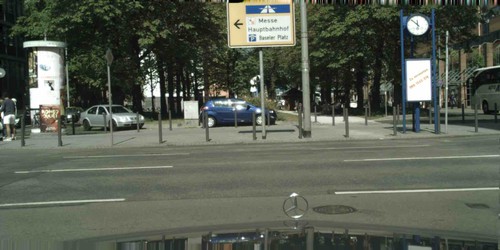} &
\includegraphics[width=0.16666666666667\linewidth]{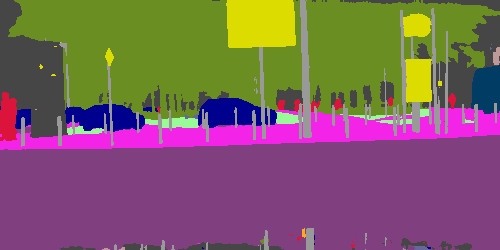} &
\includegraphics[width=0.16666666666667\linewidth]{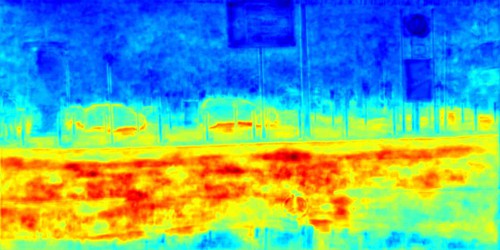} &
\includegraphics[width=0.16666666666667\linewidth]{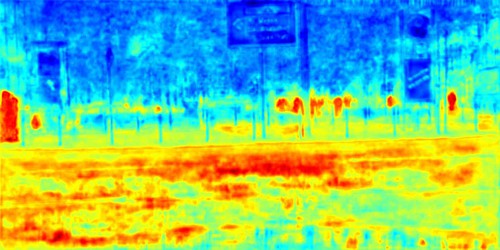} &
\includegraphics[width=0.16666666666667\linewidth]{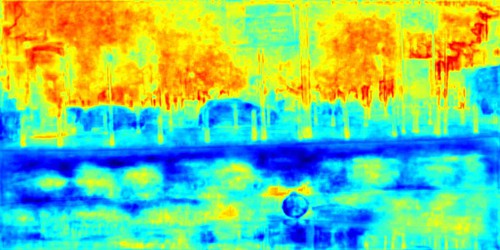} &
\includegraphics[width=0.16666666666667\linewidth]{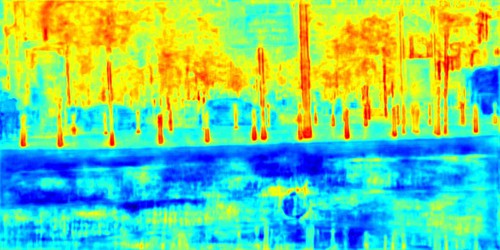} \\
Image & Prediction & Car & Person & Building & Pole \\ 
\end{tabular} \egroup 
\end{center}
\vspace{-2mm}
\caption{Visualization of guided attention for 4 general classes. Guided attention focuses on the boundary of co-occurred objects/things.}
\label{vis:general}
\vspace{-3mm}
\end{figure*}
\begin{figure*}[!t]
\begin{center}
\bgroup 
 \def\arraystretch{0.2} 
 \setlength\tabcolsep{0.2pt}
\begin{tabular}{cccc}
\includegraphics[width=0.25\linewidth]{gt/frankfurt_000000_001016_leftImg8bit.jpg} &
\includegraphics[width=0.25\linewidth]{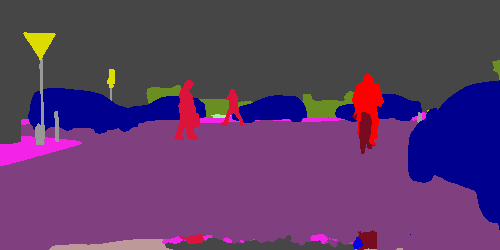} &
\includegraphics[width=0.25\linewidth]{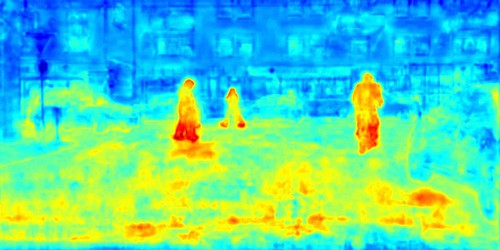} &
\includegraphics[width=0.25\linewidth]{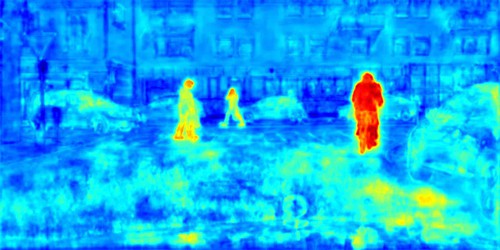}  \\
Image & Prediction & Person & Rider \\ 
\end{tabular} \egroup 
\end{center}
\vspace{-2mm}
\caption{Visualization of guided attention for Person/Rider. Guided attention is capable of differentiating semantically similar classes.}
\label{vis:similar}
\vspace{-2mm}
\end{figure*}
\begin{figure*}[!t]
\begin{center}
\bgroup 
 \def\arraystretch{0.2} 
 \setlength\tabcolsep{0.2pt}
\begin{tabular}{ccccc}
\includegraphics[width=0.2\linewidth]{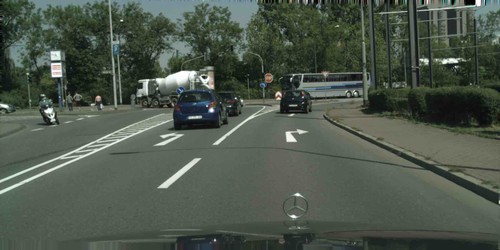} &
\includegraphics[width=0.2\linewidth]{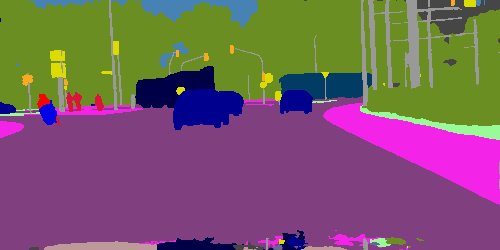} &
\includegraphics[width=0.2\linewidth]{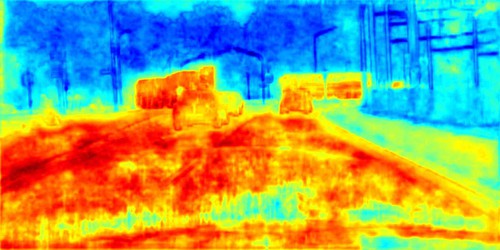} &
\includegraphics[width=0.2\linewidth]{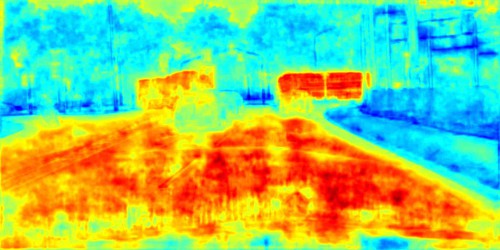} &
\includegraphics[width=0.2\linewidth]{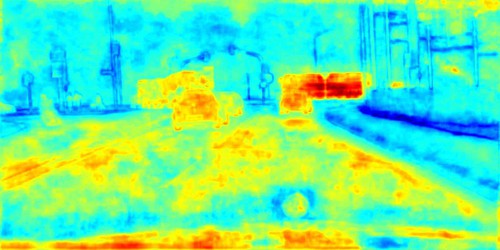}\\
\includegraphics[width=0.2\linewidth]{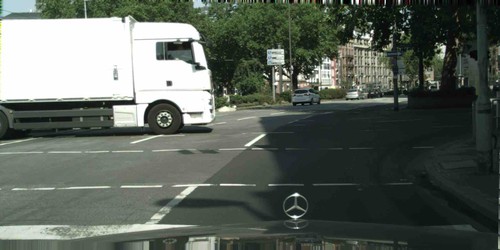} &
\includegraphics[width=0.2\linewidth]{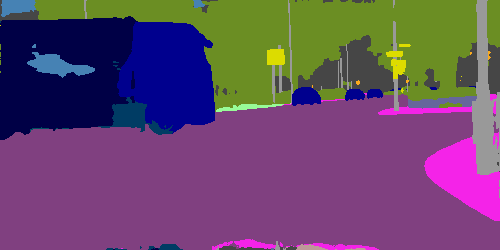} &
\includegraphics[width=0.2\linewidth]{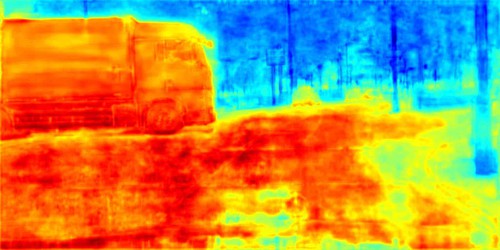} &
\includegraphics[width=0.2\linewidth]{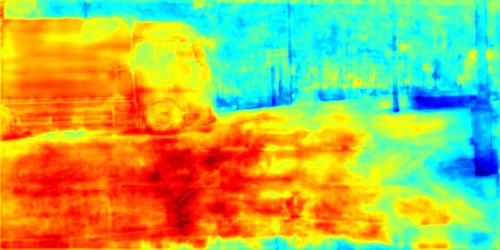} &
\includegraphics[width=0.2\linewidth]{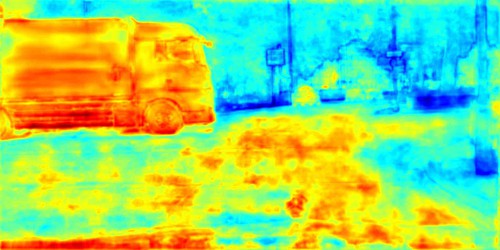}\\
Image & Prediction & Truck & Bus & Train \\ 
\end{tabular} \egroup 
\end{center}
\vspace{-2mm}
\caption{Visualization of guided attention for Truck/Bus/Train. Our failure cases where guided attention confuses among Truck/Bus/Train.}
\label{vis:failure}
\vspace{-2mm}
\end{figure*}

\subsection{Visualization of Guided Attention}\label{sec:vis}
In this section, we visualize the learned Guided Attention in our best model variant (a stack of two encoder-decoder structures with ResNet-50 as encoder backbone). The Guided Attention maps (with 256 channels) is obtained by applying a $1\times1$ convolution with sigmoid activation to the prediction in the 1st stage decoder output. Therefore, we have a convolution weight matrix with size $C \times 256$ (Figure~\ref{fig:vis_method} top-right), where $C$ is the number of semantic classes on the dataset. To visualize the attention for class $c$, we would like to know which channels among the 256 channels in the Guided Attention map that the class $c$ contributes most. Therefore, for class $c$, we extract the corresponding $1\times256$ convolution weight vector (Figure~\ref{fig:vis_method} red row in matrix) from the $C \times 256$ matrix. In the vector, we then select the indexes of the top $15$ largest weights (Figure~\ref{fig:vis_method} yellow elements in vector), which is used to index the corresponding channels in the Guided Attention maps (Figure~\ref{fig:vis_method} yellow slices from the purple Guided Attention maps), \ie, those channels in the Guided Attention maps have the largest responses for the class $c$. Then, we visualize the attention by taking $l_{2}$ norm of the selected channels.

\noindent{\bf General classes.} We visualize the learned Guided Attention for four representative classes in Figure~\ref{vis:general}. `Car' and `Person' are most common `thing' classes in the Cityscapes dataset. `Building' is a common `stuff' class and `Pole' is a common thin `stuff' in Cityscapes. The activations are normalized between 0 (blue color) and 1 (red color).

From Figure~\ref{vis:general}, we observe several interesting behaviors:
\begin{itemize}
    \setlength\itemsep{0em}
    \item 
    The guided attention learns localization of objects. The activations for `thing' align quite well with the actual position of those objects.
    \item
    Guided attention focus on object co-occurrence. For example, `Car' and `Person' objects are usually on the road and the attentions for these classes learn to focus on both corresponding instances and road.
    \item
    Guided attention can find small objects. For example, there are multiple thin `Poles' in the third row of Figure~\ref{vis:general} and guided attention can find most of them.
\end{itemize}
\noindent{\bf Semantically similar classes.}
We find guided attention is also capable of differentiating semantically similar classes. In Figure~\ref{vis:similar}, we visualize the attention for two semantically similar classes: `Person' and `Rider'. The attention for `Rider' mainly fires for the rider instance on the right, and it does not fire for the two person instances on the left of the image. Our guided attention makes the features, passed to the next stage, more discriminative to semantically similar classes through the injected supervision, allowing our SPGNet to achieve better results on both `Person' and `Rider' classes than other state-of-the-art models, as shown in Table.~\ref{table:cityscapes_sota_detailed}.

\noindent{\bf Failure cases.}
Our SPGNet confuses among `Truck', `Bus' and `Train'. We visualize the attentions for these classes in Figure~\ref{vis:failure}. We observe that the Guided Attention maps for these classes usually activate together on the same object. It potentially produces features that are less discriminative to those classes, resulting in our worse performance on `Truck', `Bus' and `Train', as shown in Table.~\ref{table:cityscapes_sota_detailed}.

\subsection{Generalization to Other Datasets}
\begin{table}[!t]\setlength{\tabcolsep}{9pt}
\centering
\scalebox{0.9}{
\begin{tabular}{c c c c}
\toprule[0.2em]
 Method & Extra data & Multi-scale & mIoU ($\%$)\\
\toprule[0.2em]
 Liang \etal~\cite{liang2017interpretable} & \xmark & \cmark & 63.57 \\
 Xia \etal~\cite{xia2017joint} & \cmark & \cmark & 64.39 \\
 Fang \etal~\cite{fang2018weakly} & \cmark & \cmark & 67.60 \\
 DPC~\cite{dpc2018} & \xmark & \cmark & 71.34 \\
\midrule
 SPGNet (Ours) & \xmark & \xmark & 67.23 \\
 SPGNet (Ours) & \xmark & \cmark & 68.36 \\
\bottomrule[0.1em]
\end{tabular}
}
\caption{ Pascal Person-Part validation set performance.}
\label{table:pascal_person_part}
\end{table}
To demonstrate that our model can be generalized to other datasets, we perform more experiments on the PASCAL VOC 2012 \cite{Everingham10} and PASCAL Person-Part \cite{chen2014detect}. For both datasets, we follow the settings in~\cite{deeplabv3plus2018} to train the model with a crop size of $513\times513$, batch size of 28 for 30,000 iterations.

\noindent\textbf{PASCAL VOC 2012}: The SPGNet with a stack of 2 ResNet-50 achieves 77.33 mIoU. The performance of SPGNet is comparable with the current state-of-the-art ResNet-101 DeepLabV3+~\cite{deeplabv3plus2018} which achieves 77.37 mIoU with encoder stride=32 for a fair comparison.

\noindent\textbf{PASCAL Person-Part}: Table~\ref{table:pascal_person_part} shows comparison with state-of-the-art results on Pascal Person-Part. Our SPGNet with a stack of 2 ResNet-50 achieves 67.23 mIoU with a single scale input, and 68.36 mIoU with multi-scale inputs. Note that our SPGNet does not require extra MPII training data~\cite{andriluka20142d}, as used in \cite{xia2017joint,fang2018weakly}.

\section{Conclusion}
We have proposed the SPGNet which demonstrates state-of-the-art performance for semantic segmentation on Cityscapes. Our proposed SPG module employs the `supervise-and-excite' framework, where the local features are reweighted via the guidance from semantic prediction. The Guided Attention maps within the SPG module allows us to visually interpret the corresponding reweighting mechanism. Our experimental results show that a two-stage encoder-decoder network paired with our SPG module can significantly outperform its one-stage counterpart with similar parameters and computations. Finally, we plan to explore a more computationally efficient encoder-decoder structure for semantic segmentation in the future.

\paragraph{Acknowledgments}
This work is in part supported by IBM-Illinois Center for Cognitive Computing Systems Research (C3SR) - a research collaboration as part of the IBM AI Horizons Network and Intelligence Advanced Research Projects Activity (IARPA) via contract D17PC00341, ARC DECRA DE190101315. The U.S. Government is authorized to reproduce and distribute reprints for Governmental purposes notwithstanding any copyright annotation thereon.  Disclaimer: The views and conclusions contained herein are those of the authors and should not be interpreted as necessarily representing the official policies or endorsements, either expressed or implied, of IARPA, DOI/IBC, or the U.S. Government. The authors thank Samuel Rota Bul{\`o} and Peter Kontschieder for the valuable discussion about the global pooling kernel size.

\appendix
\section{Extra Ablation Studies}

We provide extra ablation studies on Cityscapes val set.

\noindent{\bf Effect of upsample module.} 
\begin{table}[!t]\setlength{\tabcolsep}{9pt}
\centering
\scalebox{0.9}{
\begin{tabular}{c c c c}
\toprule[0.2em]
 Upsample Module    & mIoU ($\%$) & \#Params & \#FLOPs \\
\toprule[0.2em]
 FPN-style~\cite{lin2017feature}   & $70.01$     & $11.5$M  & $118.6$B \\
\midrule
 Ours & $\textbf{71.50}$     & $11.6$M  & $107.5$B \\
\bottomrule[0.1em]
\end{tabular}
}
\caption{Cityscapes val ablation studies on upsample module. All models use ResNet-18 in encoder. Our proposed upsample module requires fewer FLOPs and attains a better performance than the FPN-style upsample module.}
\label{table:cityscapes_upsample}
\end{table}
We perform experiments to demonstrate the effectiveness of our proposed upsample module. %
We compare the decoder equipped with our proposed upsample module against the one using FPN-style upsample module \cite{lin2017feature} (\ie, bilinear upsample + residual blocks \vs. nearest-neighbor upsample + single convolutions). In these experiments, we use ResNet-18 for encoder and we do not use global average pooling. For a fair comparison, we follow \cite{lin2017feature} to implement FPN decoder module and only use the largest resolution feature maps for prediction. We also add synchronized Inplace-ABN after all convolutions in our FPN implementation. Decoder channels are set to $128$ for both cases. Results are shown in Table~\ref{table:cityscapes_upsample}. FPN-style upsample module and our proposed module have similar parameters but our upsample module requires $10$B fewer FLOPs than the FPN-style module, thanks to the bottleneck design in residual blocks. Furthermore, using our upsample module, the performance is almost $1.5$ mIoU better than the FPN-style upsample module.

\noindent{\bf Effect of global pooling.} 
\begin{table}[!t]\setlength{\tabcolsep}{6pt}
\centering
\scalebox{0.84}{
\begin{tabular}{c c c c c}
\toprule[0.2em]
 GAP    &  Test Strategy    & mIoU ($\%$) & \#Params & \#FLOPs \\
\toprule[0.2em]
 \xmark   &   -        & $71.50$     & $11.6$M  & $107.5$B \\
\midrule
 \cmark      &   GAP     & $72.87$     & $11.7$M  & $107.6$B \\              
 \cmark      &   TILED   & $74.33$     & $11.7$M  & $8(\text{tiles})\times30.6$B \\              
 \cmark      &   AP    & $\textbf{74.48}$     & $11.7$M  & $107.6$B \\
\bottomrule[0.1em]
\end{tabular}
}
\caption{Cityscapes val ablation studies on global average pooling. All models use ResNet-18 in encoder. Adding global average pooling (GAP) is beneficial. Replacing GAP with average pooling (AP) is important during inference.}
\label{table:cityscapes_gap}
\end{table}
We experiment with the effect of Global Average Pooling (GAP) by using a single-stage encoder-decoder with ResNet-18 as encoder backbone. The GAP operation is deployed after the encoder features. %
The decoder module uses 128 channels.

We compare three strategies during inference:
\vspace{-0.2cm}
\begin{enumerate}
    \setlength\itemsep{0em}
    \item GAP: Use global average pooling during inference on the $1024\times2048$ image~\cite{liu2015parsenet}.
    \item TILED: Crop overlapping patches within the image that have the same size as training crop size (\eg $769\times769$), and use $\frac{1}{3}$ overlap between patches (\eg, overlap with 256 pixels)~\cite{zhao2017pyramid}. 
    \item AP: Replace global average pooling with an average pooling whose kernel size is the same as training crop size divided by the stride of that feature maps~\cite{porzi2019seamless}.
\end{enumerate}

\vspace{-0.2cm}
As shown in Table~\ref{table:cityscapes_gap}, we observe that using global average pooling (GAP) only improves the performance slightly by 1.3\% due to the asymmetric setting during training and inference (\ie, train with crop size $769\times769$ but inference with image size $1024\times2048$). The TILED strategy resolves this problem by employing the same pooling kernel size during training and inference. However, it introduces extra computation since it requires processing redundant pixels within the overlapped regions among patches. Furthermore, it requires some heuristics to resolve the conflicts within the overlapped regions (\eg, average the predictions in the overlapped regions), which may lead to sub-optimal merging. On the other hand, the AP strategy is more efficient than the TILED strategy and performs slightly better, since no overlapped regions are processed.

{\small
\bibliographystyle{ieee_fullname}
\bibliography{egbib}
}

\end{document}